\documentclass[conference]{IEEEtran}
\IEEEoverridecommandlockouts
\usepackage[numbers,sort&compress]{natbib}
\usepackage{mdframed} 
\usepackage{xcolor}
\usepackage[utf8]{inputenc}
\usepackage{amsmath,amssymb,amsfonts}
\usepackage{graphicx}
\usepackage{newunicodechar}
\newunicodechar{≥}{\geq}
\usepackage{url}
\usepackage{booktabs} 
\usepackage[font=small,skip=4pt]{caption}
\usepackage{enumitem}
\setlist[itemize]{leftmargin=1.2em,topsep=2pt,itemsep=1pt,parsep=0pt}
\setlist[enumerate]{leftmargin=1.4em,topsep=2pt,itemsep=1pt,parsep=0pt}
\mdfdefinestyle{promptbox}{%
  backgroundcolor=gray!10,
  linecolor=gray!50,
  linewidth=0.4pt,
  roundcorner=3pt,
  innertopmargin=2mm,
  innerbottommargin=2mm,
  innerleftmargin=2mm,
  innerrightmargin=2mm,
  frametitle={Grounding Prompt},
  frametitlefont=\bfseries\small
}

\begin{document}

\title{Why Does Grounding Hurt Medical VQA? Benchmarking, Diagnosis, and Fine-Tuning of Vision-Language Models}

\author{
\IEEEauthorblockN{%
Xupeng Chen\textsuperscript{*,1},\quad
Binbin Shi\textsuperscript{2},\quad
Chenqian Le\textsuperscript{1},\quad
Jun Li\textsuperscript{1},\quad
Qifu Yin\textsuperscript{3},\\[3pt]
Lang Lin\textsuperscript{3},\quad
Haowei Ni\textsuperscript{3},\quad
Ran Gong\textsuperscript{1},\quad
Panfeng Li\textsuperscript{4}}\\[2pt]
\IEEEauthorblockA{%
\textsuperscript{1}\textit{New York University, New York, USA}\quad
\textsuperscript{2}\textit{Tsinghua University, Beijing, China}\\[1pt]
\textsuperscript{3}\textit{Columbia University, New York, USA}\quad
\textsuperscript{4}\textit{University of Michigan, Ann Arbor, USA}\\[1pt]
\textsuperscript{*}Corresponding author: \texttt{xc1490@nyu.edu}}
}
\maketitle

\begin{abstract}
Vision-language models (VLMs) are increasingly applied to medical visual question answering (Med-VQA), yet whether they can \emph{localize} the evidence behind their answers---a prerequisite for clinical auditability---is poorly characterized. We separately evaluate VQA reasoning and visual grounding for four recent frontier VLMs (GPT-5.1, GPT-5.5, Gemini-2.5-Pro, Gemini-3-Flash), two domain-specific medical VLMs (Lingshu, MedGemma), and a dedicated open-vocabulary detector (Grounding DINO) on VQA-RAD and SLAKE. Two findings challenge the intuition that ``add grounding to improve VQA.'' First, \textbf{no model localizes medical targets well}: every off-the-shelf system---frontier, medical-specialized, or dedicated detector---scores mean IoU 0.05--0.24 on our SLAKE grounding split, at or barely above a trivial center-box baseline (0.10), with Acc@0.5 below 20\%. Second, and counter to the common ``localize-then-answer'' paradigm, \textbf{cropping to a bounding box degrades VQA even when the box is a perfect oracle}: on the matched subset where oracle ground-truth boxes are applied, GT-grounding \emph{lowers} closed-ended accuracy for every model (by 0.9--18.0 points versus using the full image)---consistent with the crop discarding global context the model relies on. Because the oracle box removes localization error by construction, the problem is not that perception is a recoverable bottleneck, but that grounding-by-cropping is itself the wrong interface. Finally, we show constructively that the two channels need not conflict: supervised fine-tuning of Qwen-2.5-VL-7B on answers \emph{alone} silently destroys box-evidence emission (0/418 parseable boxes), whereas mixing in a small amount of grounding supervision restores localization to 0.36 IoU---above every zero-shot model---while preserving answer accuracy.
\end{abstract}


\section{Introduction}
Large Vision-Language Models (VLMs) have shown remarkable zero-shot capability on general visual content, raising hopes that they could assist clinicians in interpreting medical images such as X-rays and CT scans. Yet Medical Visual Question Answering (VQA) demands two equally critical skills: high-level reasoning, at which modern LLMs excel, and low-level perception---the precise localization of anatomical structures or pathologies in a complex medical image. A failure of perception can render even the most sophisticated reasoning useless.

A natural hypothesis is that many apparent `reasoning' failures in medical VQA are really failures of visual grounding---the model looks at the wrong region---and that supplying the correct region would fix them. We test this hypothesis directly and find that it does not hold: models localize poorly, and even a \emph{perfect} region, supplied by cropping to an oracle box, does not help and usually hurts. This reframes the problem from ``perception is a recoverable bottleneck'' to ``grounding-by-cropping is the wrong interface, and answer-only training does not keep the evidence channel intact.'' Our key contributions are:

\begin{itemize}
    \item We benchmark medical visual grounding across a broad panel---four frontier VLMs, two medical-specialized VLMs, and a dedicated open-vocabulary detector---and show that \emph{none} localizes medical targets well (mean IoU 0.05--0.24, at or barely above a trivial center-box baseline), so poor localization is universal rather than a quirk of any one model family.
    \item We show that self-grounding (the same model localizes then answers) lowers open-ended recall for every model, and---critically---that \textbf{even oracle GT-box cropping degrades VQA} on the matched subset (closed-ended accuracy drops 0.9--18.0 points versus the full image), consistent with the crop discarding global context. This overturns the intuition that accurate grounding recovers VQA.
    \item Through a matched four-arm fine-tuning study, we show answer-only supervised fine-tuning silently eliminates box-evidence emission (0/418 parseable boxes), that format rehearsal alone restores boxes but not localization (IoU 0.02), and that adding a small amount of grounding supervision restores localization to 0.36 IoU---above every zero-shot model---while preserving answer accuracy. The answer and evidence channels are separable, and the loss is a fixable recipe choice, not an inherent trade-off.
\end{itemize}

\begin{figure*}[t!]
    \centering
    \includegraphics[width=0.95\textwidth]{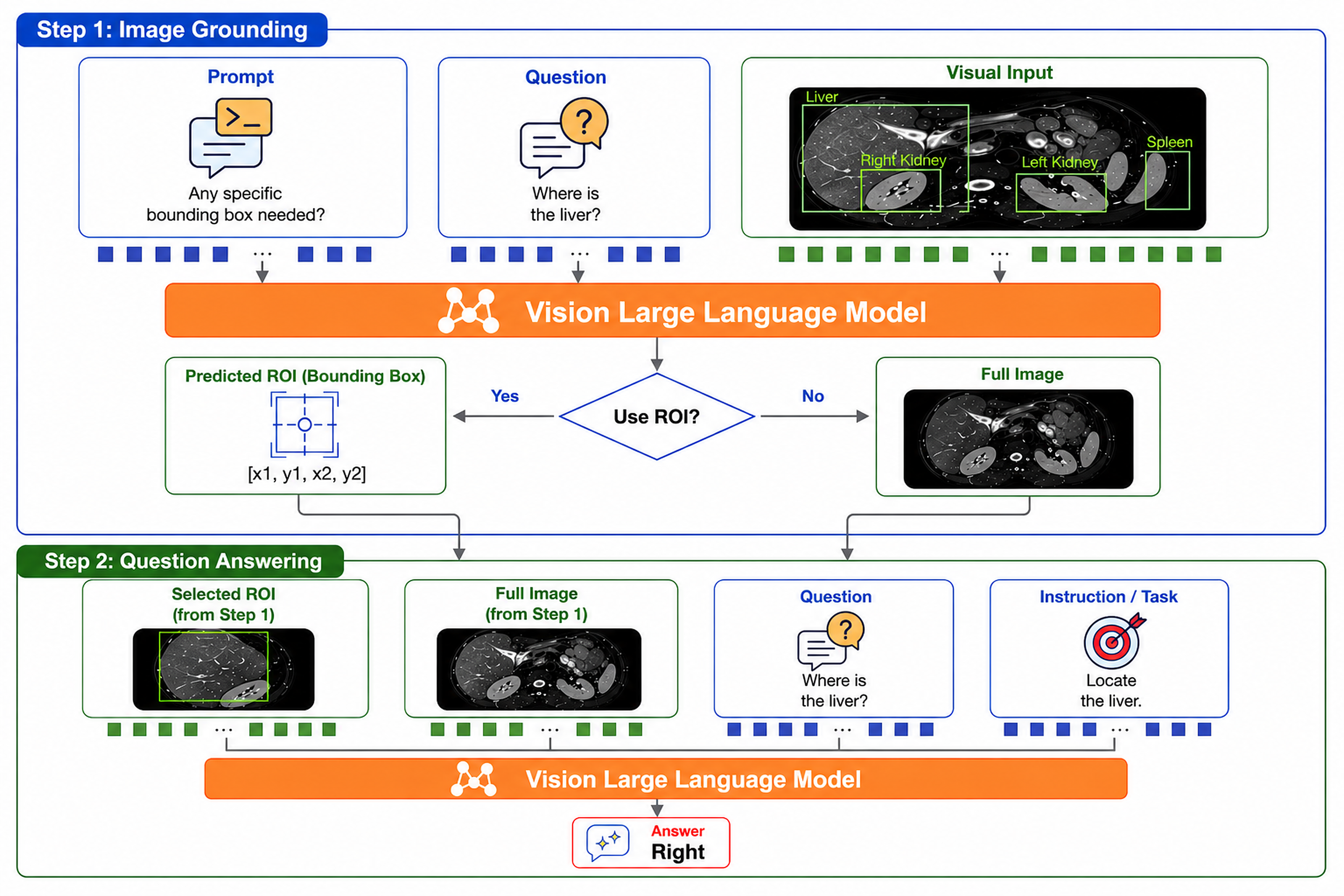}
    \caption{\textbf{The two-step grounding-enhanced VQA pipeline.} Step~1: the VLM jointly consumes the prompt, question, and medical image and emits a predicted ROI $[x_1,y_1,x_2,y_2]$; depending on the configuration, the ROI is used to crop the image (Self-/GT-Grounding) or the full image is forwarded directly (Direct VQA). Step~2: the same VLM is re-invoked with the selected input, the question, and a task instruction to produce the final answer. We evaluate three configurations---Direct, Self-Grounding, and oracle GT-Grounding VQA---and find that the cropping step itself degrades accuracy, even when the box is a perfect oracle.}
    \label{fig:pipeline}
\end{figure*}

\section{Related Work}
\paragraph{Large Vision-Language Models.} Foundational contrastive models such as CLIP~\cite{radford2021clip} and BLIP~\cite{li2022blip} established vision--language alignment; modern frontier VLMs (Gemini~\cite{comanici2025gemini}, the GPT series~\cite{openai2025gpt5,hurst2024gpt4o}, InternVL~\cite{chen2024internvl}) and strong open-source systems (GLM~\cite{vteam2025glm45vglm41vthinkingversatilemultimodal}, Qwen-VL~\cite{bai2025qwen25vltechnicalreport}, LLaVA~\cite{liu2023llava}) extend this paradigm with LLM-based reasoning. Their reliability in high-stakes specialized domains such as medicine, however, remains underexplored.

\paragraph{Medical Visual Question Answering.} VQA-RAD~\cite{lau2018vqarad}, SLAKE~\cite{liu2021slake}, and PathVQA~\cite{he2020pathvqa} are the standard Med-VQA benchmarks; SLAKE is the one that provides organ-level bounding boxes, which is why our grounding analysis is built on it (Section~\ref{subsec:datasets}). Concurrent to our work, Liu et al.~\cite{liu2025howfar} benchmark medical VLMs across eight datasets, separating understanding from reasoning; our study is complementary, asking not how well models answer but whether the grounding interface itself helps or hurts.

\paragraph{Medical VLM Fine-Tuning.} Because zero-shot generalists trail in-domain specialists on medical VQA, several works adapt VLMs to medicine via supervised fine-tuning. LLaVA-Med~\cite{li2023llavamed} trains LLaVA in two stages using PubMed figure--caption pairs and downstream Med-VQA instruction data. Our fine-tuning study complements this line of work by (i) using a more recent base (Qwen~2.5~VL~7B) and (ii) relying purely on the standard train splits of the three benchmarks as a transparent and reproducible recipe.

\paragraph{Visual Grounding in VLMs.} Specialized models such as Grounding DINO~\cite{liu2024groundingdinomarryingdino} and SAM~\cite{kirillov2023segment} achieve strong open-set localization and segmentation. Some generalist VLMs (e.g., GLM-4V-Plus) advertise built-in grounding, but their precise-localization ability in specialized domains like medicine has not been systematically benchmarked against their reasoning performance---the gap our work aims to fill.

\section{Experiments}
To rigorously evaluate the reasoning and grounding capabilities of current leading Vision-Language Models (VLMs), we designed a comprehensive benchmark consisting of two distinct tasks: a standard Medical VQA evaluation and a precise visual grounding evaluation. This section details the models, datasets, evaluation metrics, and experimental setup used in our study.

\subsection{Models}
Our panel spans three families. \textbf{Frontier VLMs}: GPT-5.1, GPT-5.5~\cite{openai2025gpt5}, Gemini-2.5-Pro, Gemini-3-Flash~\cite{comanici2025gemini}, and---for the answer comparison---GLM-4V-Plus~\cite{vteam2025glm45vglm41vthinkingversatilemultimodal} and Qwen-VL-Max~\cite{bai2025qwen25vltechnicalreport}; GLM-4V-Plus is notable as the only one advertising built-in visual grounding. \textbf{Medical-specialized VLMs}: Lingshu-7B and MedGemma-4B, used in the grounding benchmark to test whether domain pre-training buys localization. \textbf{A dedicated detector}: Grounding DINO~\cite{liu2024groundingdinomarryingdino}, an open-vocabulary text-conditioned localizer, as a non-VLM reference point for the grounding task. Exact API model identifiers and access dates are listed in Appendix~\ref{app:models}.

\subsection{Datasets}
\label{subsec:datasets}
We use three standard Med-VQA benchmarks, covering both radiology and pathology. Across all three we follow the common convention of categorizing questions as either \emph{closed-ended} (yes/no or single-word, e.g., ``are regions of the brain infarcted?'' $\to$ ``yes'') or \emph{open-ended} (descriptive, e.g., ``what abnormality is seen?'' $\to$ ``blind-ending loop of bowel arising from the cecum'').

\textbf{VQA-RAD}~\cite{lau2018vqarad} contains 315 radiology images and 3,515 clinician-curated QA pairs. We evaluate on the standard 451-QA test split. VQA-RAD does \emph{not} provide bounding-box annotations, so we use it only for the VQA task and it is excluded from the grounding and GT-Ground experiments.

\textbf{SLAKE}~\cite{liu2021slake} is a semantically labeled, knowledge-enhanced radiology dataset with 642 images, over 14,000 QA pairs, and rich spatial annotations (segmentation masks and bounding boxes for organs and abnormalities). We use SLAKE \emph{twice}: (i) its 1{,}061-QA test split for VQA evaluation, and (ii) a grounding split that we programmatically derive from SLAKE's organ-level bounding boxes by generating an explicit grounding query (e.g., ``Locate the liver.'') for each annotated object, yielding 418 image--query--bbox triples for the grounding benchmark and for the GT-Ground VQA condition.

We do not evaluate on PathVQA~\cite{he2020pathvqa}: it provides no bounding-box annotations, so it cannot support either the grounding benchmark or the GT-grounding condition that this study is built around.

\subsection{Evaluation Metrics}
\textbf{VQA.} We report closed-ended accuracy and open-ended recall. Ground-truth answers are routed into the two types by a keyword vocabulary, then matched with a rule-based pipeline tailored to medical language: closed-ended answers are first screened for direct contradictions (e.g., ``left'' vs.\ ``right'') against a curated dictionary, then matched through a synonym map; open-ended answers use synonym-expanded token overlap with a dynamic threshold (50\% for answers $\leq$3 words, 30\% otherwise) and a laterality contradiction penalty. Because this matching is heuristic rather than human-validated, we discuss its limitations in Section~\ref{sec:limitations}.

\textbf{Grounding.} We report the mean Intersection over Union (IoU), the Dice coefficient, and Accuracy@0.5 (the fraction of predictions with IoU$\geq$0.5) between the predicted and ground-truth bounding boxes.

\subsection{Experimental Setup}
Our evaluation methodology is centered around a two-step pipeline (Figure~\ref{fig:pipeline}), applied consistently across all models. Proprietary models are accessed via their official APIs; open-source models are run in a controlled local environment.

\begin{enumerate}
    \item \textbf{Step 1 -- Grounding.} The model is prompted as an ``expert radiologist AI'' to return the bounding box of the object referenced in the question, in a strict \texttt{[x1, y1, x2, y2]} format, with the original image width and height included in the prompt. A regular expression extracts the first tuple from the response; unparseable outputs are assigned a sentinel box $[0,0,0,0]$ and kept in the final score so that reported numbers reflect real-deployment robustness.

    \item \textbf{Step 2 -- Question Answering.} The model, re-prompted as a ``clinical Q\&A assistant,'' answers the original question using the visual input from Step 1: either the cropped region (if grounding was used) or the full image (otherwise). The prompt asks for brief answers to closed questions and descriptive answers to open ones.
\end{enumerate}
Prompt, parsing, and scoring are identical across all models.

\subsection{Supervised Fine-Tuning Setup}
\label{subsec:sft}
To test how fine-tuning affects the answer and evidence channels, we fine-tune \textbf{Qwen2.5-VL-7B-Instruct} on the combined train splits of VQA-RAD and SLAKE (7{,}765 examples; 5{,}972 SLAKE and 1{,}793 VQA-RAD), with no external instruction data. Arms B and C add grounding-formatted prompts on top of this identical answer set, differing only in whether the supplied boxes are shuffled (B) or the true annotations (C, a 15\% mix); the matched zero-shot base is the untouched checkpoint evaluated through the same harness. We train all parameters (vision encoder and language model) for \textbf{3 epochs} with \textbf{effective batch size 32} (per-device batch 4, gradient accumulation 8), AdamW with learning rate \textbf{1e-5}, cosine decay, and warmup ratio 0.03. Full fine-tuning uses DeepSpeed ZeRO-3 on \textbf{2$\times$ NVIDIA H200}; the LoRA variant (rank 16, vision tower frozen) runs on \textbf{1$\times$ H200}. We evaluate the fine-tuned model on the same test splits used for zero-shot evaluation and report closed-ended accuracy and open-ended recall.

\section{Comparative Analysis of Model Performance}
\subsection{Zero-shot Evaluation of Open-Source and Proprietary VLMs}

Before turning to grounding, Table~\ref{tab:vqa_comparison} situates answer accuracy on the two benchmarks we ran ourselves, VQA-RAD and SLAKE. All of our cells are full-$N$ runs with no empty responses; frontier cells come from reruns performed after we found that an earlier evaluation had recorded quota-failed, empty responses (Section~\ref{sec:limitations}).

\begin{table*}[t!]
    \centering
    \caption{Open-ended recall (Open) and closed-ended accuracy (Closed), in \%, under our heuristic answer scorer. The first two blocks are \emph{our own} evaluations at full $N$: frontier API models, then the open-weight Qwen2.5-VL-7B base and our four fine-tuning arms. The third block reproduces numbers as \emph{reported by prior work}; those systems use different evaluation harnesses and answer scorers, so we deliberately do not bold a ``winner'' across blocks and make no state-of-the-art claim. We report VQA-RAD and SLAKE only, the two benchmarks we ran ourselves.}
    \label{tab:vqa_comparison}
    \small 
    \setlength{\tabcolsep}{4pt} 
    \renewcommand{\arraystretch}{1.2}
    \begin{tabular}{@{}l cc cc@{}}
    \toprule
    & \multicolumn{2}{c}{\bf VQA-RAD ($N{=}451$)} & \multicolumn{2}{c}{\bf SLAKE ($N{=}1061$)} \\
    \cmidrule(r){2-3} \cmidrule(l){4-5}
    Method & Open & Closed & Open & Closed \\
    \midrule
    \multicolumn{5}{@{}l}{\it Zero-shot frontier VLMs (our evaluation, full-$N$ reruns)} \\
    GPT-5.5           & 46.5 & 79.4 & 47.6 & 80.1 \\
    Gemini-3-Flash    & 49.0 & 75.0 & 47.8 & 75.1 \\
    Gemini-2.5-Pro~\cite{comanici2025gemini}    & 45.8 & 73.0 & 35.4 & 70.0 \\
    GPT-5.1~\cite{openai2025gpt5}           & 34.8 & 67.6 & 40.6 & 58.1 \\
    GLM-4V-Plus~\cite{vteam2025glm45vglm41vthinkingversatilemultimodal}          & 48.4 & 53.4 & 47.6 & 49.7 \\
    Qwen-VL-Max~\cite{bai2025qwen25vltechnicalreport}       & 41.9 & 70.6 & 51.2 & 66.9 \\
    \midrule
    \multicolumn{5}{@{}l}{\it Open-weight base and our fine-tuning arms (Sec.~\ref{subsec:sft_results})} \\
    Qwen2.5-VL-7B (zero-shot base) & 38.1 & 66.2 & 37.2 & 64.6 \\
    \quad + A. answer-only SFT     & 29.0 & 66.6 & 80.0 & 77.1 \\
    \quad + B. format rehearsal    & 29.7 & 67.2 & 80.3 & 79.7 \\
    \quad + C. grounding mix       & 29.0 & 65.2 & 83.0 & 78.4 \\
    \midrule
    \multicolumn{5}{@{}l}{\it Reported by prior work (different harnesses; not directly comparable)} \\
    LLaVA~\cite{liu2023llava}  & 50.00 & 65.07 & 78.18 & 63.22 \\
    LLaVA-Med~\cite{li2023llavamed} & 61.52 & 84.19 & 83.08 & 85.34 \\
    VL Encoder--Decoder~\cite{bazi2023vision} & 71.49 & 82.47 &  \multicolumn{2}{c}{--} \\
    Q2ATransformer~\cite{liu2023q2atransformer} & 79.19 & 81.20 &  \multicolumn{2}{c}{--} \\
    Prefix T.\ Medical LM~\cite{van2023open} & \multicolumn{2}{c}{--} & 84.30 & 82.01 \\
    PubMedCLIP~\cite{eslami2023pubmedclip} & 60.10 & 80.00 & 78.40 & 82.50 \\
    BiomedCLIP~\cite{zhang2023large} & 67.60 & 79.80 & 82.05 & 89.70 \\
    \bottomrule
    \end{tabular}
\end{table*}

Two patterns matter for what follows. (i) Frontier models answer respectably but are not saturated (best SLAKE closed 80.1\%, best open 47.8\%), and---read against the reported specialist numbers, with the harness caveat above---in-domain supervision remains competitive. (ii) In-domain fine-tuning of a 7B open-weight model produces a large SLAKE gain (closed 64.6$\to$77.1--79.7\%, open 37.2$\to$80.0--83.0\%) but \emph{lowers} VQA-RAD open recall (38.1$\to$29.0--29.7\%): our training mix is SLAKE-heavy, so the model specializes to SLAKE's answer style at the cost of the out-of-mix benchmark. We report this rather than tuning it away, since it bounds how far the SFT result should be read---and, as Section~\ref{subsec:sft_results} shows, the answer score alone conceals a more consequential change in the model's grounding behavior.

\subsection{Grounding Accuracy of VLMs}
\label{subsec:grounding}

To assess perception directly, we benchmark a broad panel on our SLAKE-derived grounding split ($N{=}418$): four frontier VLMs, two medical-specialized VLMs (Lingshu-7B, MedGemma-4B), and a dedicated open-vocabulary detector (Grounding DINO). Table~\ref{tab:grounding} reports IoU (over parseable boxes) and Acc@0.5 against two trivial baselines---a fixed center box and the open-weight Qwen-2.5-VL-7B backbone---and, for reference, our best grounding-supervised fine-tune (Section~\ref{subsec:sft_results}).

\begin{table}[t!]
    \centering
    \caption{Visual grounding on the SLAKE split ($N{=}418$). IoU is the mean over parseable predictions; Acc@0.5 is the fraction with IoU $\geq 0.5$. No off-the-shelf model---frontier, medical-specialized, or dedicated detector---meaningfully clears the trivial center-box baseline. Only grounding-supervised fine-tuning (bottom) does.}
    \label{tab:grounding}
    \small
    \setlength{\tabcolsep}{5pt}
    \renewcommand{\arraystretch}{1.1}
    \begin{tabular}{@{}l cc@{}}
    \toprule
    Model & IoU $\uparrow$ & Acc@0.5 $\uparrow$ \\
    \midrule
    \multicolumn{3}{@{}l}{\it Frontier VLMs} \\
    GPT-5.5          & \textbf{0.24} & 17.9 \\
    Gemini-3-Flash   & 0.20 & \textbf{18.2} \\
    GPT-5.1          & 0.16 & 6.9 \\
    GLM-4V-Plus         & 0.13 & 4.3 \\
    Gemini-2.5-Pro   & 0.11 & 3.6 \\
    \midrule
    \multicolumn{3}{@{}l}{\it Medical-specialized VLMs} \\
    Lingshu-7B       & 0.07 & 2.6 \\
    MedGemma-4B      & 0.05 & 0.0 \\
    \midrule
    \multicolumn{3}{@{}l}{\it Dedicated detector} \\
    Grounding DINO   & 0.08 & 0.7 \\
    \midrule
    \multicolumn{3}{@{}l}{\it Trivial baselines} \\
    Center box       & 0.10 & 2.6 \\
    Qwen-2.5-VL-7B (base) & 0.11 & 4.1 \\
    \midrule
    \multicolumn{3}{@{}l}{\it Grounding-supervised fine-tune (ours, Sec.~\ref{subsec:sft_results})} \\
    Qwen-2.5-VL-7B + grounding mix & \textbf{0.39} & \textbf{40.2} \\
    \bottomrule
    \end{tabular}
\end{table}

The panel exposes a stark and \emph{universal} perceptual gap. The best off-the-shelf system, GPT-5.5, reaches only 0.24 IoU and 17.9\% Acc@0.5---fewer than one in five predicted boxes overlap the ground truth by half---and every other model is worse. Crucially, this is not a frontier-model quirk: the medical-specialized VLMs localize \emph{worse} (Lingshu 0.07, MedGemma 0.05), and even a dedicated open-vocabulary detector (Grounding DINO, 0.08) fails, because it is trained on natural images and anatomy is out of distribution. All of these sit at or below a trivial center box (0.10). Only fine-tuning with explicit box supervision (0.39) clears the baseline meaningfully. Medical grounding, in short, is an unsolved, distribution-specific problem that neither scale nor generic detection solves.

\begin{figure*}[t!]
    \centering
    \includegraphics[width=0.82\textwidth]{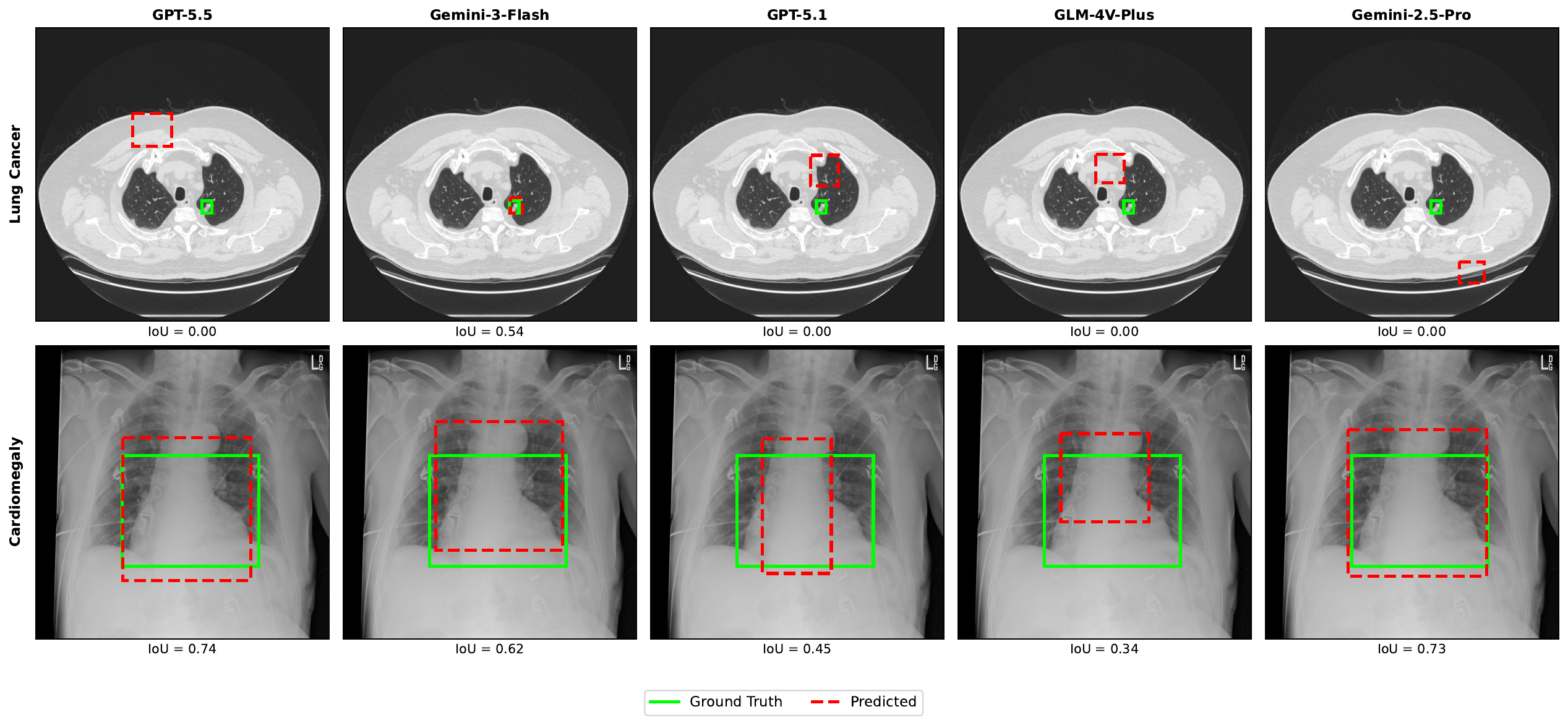}
    \caption{Qualitative grounding on two samples, illustrating the two error modes. \textbf{Top}: a small lung-cancer lesion in a CT scan---four of five models miss it entirely (IoU~=~0.00), placing the box in unrelated tissue or the wrong lung; only Gemini-3-Flash overlaps it (0.54). \textbf{Bottom}: cardiomegaly in a chest X-ray, a large target---every model overlaps the region (IoU~0.34--0.74) but oversizes or offsets the box. Together these show the scale dependence quantified in Table~\ref{tab:grounding}: performance is not uniformly poor, it collapses on small targets. Green: ground truth; red dashed: model prediction.}
    \label{fig:grounding_examples}
\end{figure*}

\textbf{Format compliance is not the bottleneck.} Low IoU is genuine localization error, not unparseable output: on the dedicated grounding task the frontier models emit a valid box on $99$--$100\%$ of queries (after a parser fix that also accepts bare coordinates), so restricting to parseable predictions leaves IoU essentially unchanged. Two systematic error modes recur where boxes \emph{are} produced, both visible in Figure~\ref{fig:grounding_examples}: \textbf{scale mismatch}---small targets (lesions, nodules) draw oversized boxes covering a whole organ or quadrant, and every model fails outright on the small lung-cancer lesion (top row)---and, for a subset of models, \textbf{laterality confusion} on mirrored chest radiographs, a clinically dangerous error; we report per-model laterality rates in Appendix~\ref{app:laterality}, as they are model-specific rather than universal.

\subsection{Does Grounding Improve VQA?}

We compare three settings: (1)~\textbf{Direct VQA}, full image + question in one step; (2)~\textbf{Self-Grounding VQA}, where the \emph{same} model first localizes then answers from the crop; and (3)~\textbf{GT-Grounding VQA}, where SLAKE's organ-level ground-truth bounding boxes provide an \emph{oracle} crop---the best case the two-step pipeline could hope for. Under GT-Grounding we crop only the anatomy-focused questions (Organ, Abnormality, Position, Size; 589 of 1{,}061 SLAKE questions); global questions (Modality, Knowledge, Plane, etc.) keep the full image. Because only that subset is actually cropped, the fair test of whether an oracle box helps is a \emph{matched} comparison on those 589 questions, which we report alongside the full-set numbers in Table~\ref{tab:grounding_vqa}.

\begin{table*}[t!]
    \centering
    \caption{Direct VQA vs.\ Self-Grounding vs.\ GT-Grounding VQA. Open-ended recall and closed-ended accuracy (\%) are reported. Self-Grounding uses the same model for both localization and answering; GT-Grounding uses ground-truth bounding boxes from SLAKE annotations for selective cropping. VQA-RAD lacks bounding-box annotations and therefore has no GT-Grounding column.}
    \label{tab:grounding_vqa}
    \small
    \setlength{\tabcolsep}{3.5pt}
    \renewcommand{\arraystretch}{1.1}
    \begin{tabular}{@{}l cc cc cc cc cc@{}}
    \toprule
    & \multicolumn{4}{c}{\bf VQA-RAD ($N{=}451$)} & \multicolumn{6}{c}{\bf SLAKE ($N{=}1061$)} \\
    \cmidrule(r){2-5} \cmidrule(l){6-11}
    & \multicolumn{2}{c}{Direct} & \multicolumn{2}{c}{Self-Ground} & \multicolumn{2}{c}{Direct} & \multicolumn{2}{c}{Self-Ground} & \multicolumn{2}{c}{GT-Ground} \\
    \cmidrule(r){2-3} \cmidrule(r){4-5} \cmidrule(r){6-7} \cmidrule(r){8-9} \cmidrule(l){10-11}
    Model & Open & Closed & Open & Closed & Open & Closed & Open & Closed & Open & Closed \\
    \midrule
    GPT-5.1          & 34.8 & 67.6 & 29.7 & 61.5 & 40.6 & 58.1 & 37.7 & 59.3 & 37.2 & 57.9 \\
    GPT-5.5          & 46.5 & \textbf{79.4} & 43.2 & \textbf{70.3} & 47.6 & \textbf{80.1} & 46.6 & \textbf{70.6} & 44.0 & \textbf{68.6} \\
    Gemini-2.5-Pro   & 45.8 & 73.0 & 36.8 & 64.9 & 35.4 & 70.0 & 34.2 & 61.6 & 36.6 & 63.9 \\
    Gemini-3-Flash   & \textbf{49.0} & 75.0 & 41.3 & 66.2 & \textbf{47.8} & 75.1 & 40.9 & 63.1 & 43.7 & 66.7 \\
    \bottomrule
    \end{tabular}
\end{table*}

\paragraph{Self-grounding degrades VQA.} Self-grounding---where the same model localizes then answers from its own crop---\emph{degrades} VQA for all four models. On SLAKE, open-ended recall falls for all four (e.g.\ Gemini-3 47.8\%$\to$40.9\%, GPT-5.1 40.6\%$\to$37.7\%), and closed-ended accuracy falls for three of the four (GPT-5.5 80.1\%$\to$70.6\%, Gemini-3 75.1\%$\to$63.1\%, Gemini-2.5 70.0\%$\to$61.6\%), with GPT-5.1 the lone exception (58.1\%$\to$59.3\%). VQA-RAD shows the same pattern (e.g.\ Gemini-3 75.0\%$\to$66.2\% closed). Two factors compound: the model's own boxes are inaccurate (Section~\ref{subsec:grounding}), and the two-step prompt is harder to constrain than a single answer. We note that a bounding-box parser that accepted only bracketed coordinates silently zeroed a large fraction of predictions in an earlier run (up to 50\% of GPT-5.1 self-grounding crops), which would make self-grounding masquerade as direct VQA; all numbers here use a corrected parser that also accepts bare coordinates, and every self-grounding run emits a valid crop.

\begin{figure}[t!]
    \centering
    \includegraphics[width=0.98\columnwidth]{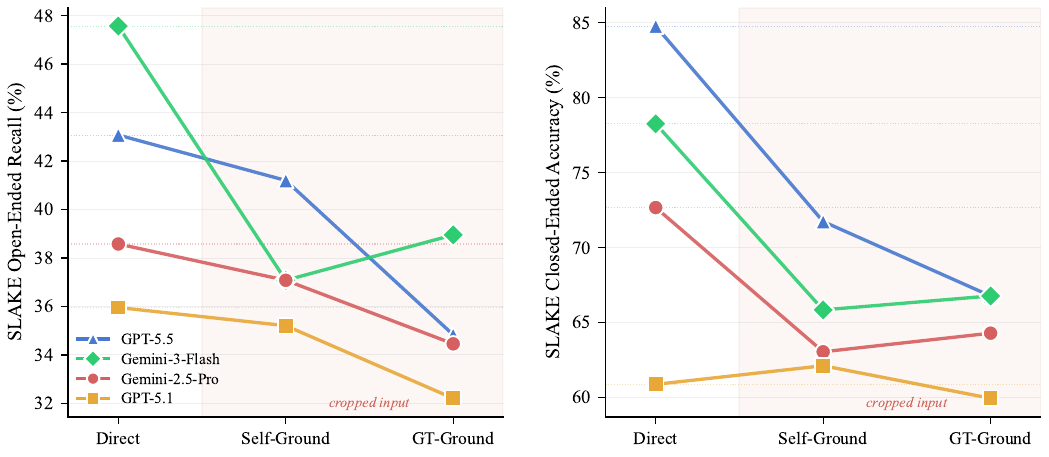}
    \caption{SLAKE VQA on the \emph{matched} anatomy subset ($n{=}589$, the questions where cropping is actually applied), scored on identical ids across all three conditions. Dotted lines mark each model's Direct (full-image) level. Oracle GT-Grounding falls below Direct for \emph{every} model on \emph{both} metrics, and self-grounding falls below Direct for every model on open-ended recall (and for three of four on closed-ended accuracy; GPT-5.1 is the exception, $+1.2$). Cropping to a box---even a perfect one---costs more than the localization it supplies.}
    \label{fig:vqa_comparison}
\end{figure}

\paragraph{Even oracle grounding hurts.} If poor localization were the bottleneck, replacing the model's boxes with SLAKE's \emph{ground-truth} boxes should recover---and ideally surpass---direct VQA. It does not. On the matched subset of 589 anatomy questions where the oracle box is actually applied, GT-grounding \emph{lowers} closed-ended accuracy for \textbf{every} model relative to using the full image: GPT-5.5 84.8\%$\to$66.8\% ($-$18.0), Gemini-3 78.3\%$\to$66.8\% ($-$11.5), Gemini-2.5 72.7\%$\to$64.3\% ($-$8.4), and GPT-5.1 60.9\%$\to$59.9\% ($-$0.9); open-ended recall drops for all four as well ($-$3.7 to $-$8.6). A \emph{perfect} region of interest, supplied for free, makes the models answer \emph{worse}. The reason is that cropping to the box discards the surrounding anatomy and global context that medical VQA relies on---modality, laterality, and organ relationships are often read from parts of the image the oracle box excludes. This directly refutes the ``perception is a recoverable bottleneck'' reading: the failure is not that the model looks in the wrong place, but that forcing it to look \emph{only} at one place, even the right one, throws away information. Grounding-by-cropping is the wrong interface for a task that is inherently global. We attribute the loss to discarded context rather than to the crop's smaller field of view on the strength of the oracle design itself---the box is correct by construction, so localization error cannot explain the drop---but a direct test that varies the crop's context window while holding the box fixed would isolate the two, and we flag it as the natural follow-up (Section~\ref{sec:limitations}).

\paragraph{Interpretation: the interface, not the perception, is the bottleneck.} Putting the two conditions together rules out the intuitive story. Self-grounding could fail for two reasons---bad boxes or a lossy pipeline---and by itself does not distinguish them. The oracle condition isolates the pipeline: with localization error removed entirely, cropping still hurts. So the dominant failure is not that models cannot find the region (though they cannot---Section~\ref{subsec:grounding}), but that the localize-then-crop-then-answer interface destroys information even when localization is perfect. Better boxes would not rescue this pipeline; a different interface is needed---one that highlights or attends to a region without discarding the rest of the image. We treat the specific magnitudes as observational (four models, one benchmark, a heuristic answer scorer), but the direction is unambiguous and consistent across every model and both metrics.

\subsection{Fine-Tuning: Answers Without Evidence, and How to Keep Both}
\label{subsec:sft_results}
Does supervised fine-tuning (SFT) on medical VQA improve grounding along with answers? We fine-tune Qwen-2.5-VL-7B and evaluate the result on \emph{both} channels, using the untouched 7B as a matched baseline (grounding IoU$\mid$valid $0.111$, SLAKE $37.2/64.6$). A four-arm design isolates the mechanism (Table~\ref{tab:sft_arms}). \textbf{(A) Answer-only} SFT raises the SLAKE answer score sharply ($80.0/77.1$; $83.0/80.1$ at a second seed) but \emph{eliminates} box evidence: the fine-tuned model emits \textbf{no parseable box on any of the 418 grounding queries} ($0/418$), answering the grounding prompt with coarse text (``Left Lung''). This collapse replicates across seeds and is invisible to any answer-only leaderboard. \textbf{(B) Adding format rehearsal} (the same grounding prompts, but with \emph{shuffled} boxes carrying no spatial signal) restores box emission ($418/418$) yet the boxes are useless (IoU $0.021$, below the base): emitting a box is not localizing. \textbf{(C) Adding real grounding supervision} (a small 15\% mix of GT-box examples) restores emission \emph{and} lifts localization to \textbf{IoU $0.390$} ($0.360$ on never-seen images---above every zero-shot model in this study, including the much larger API systems) while keeping the full answer gain ($83.0/78.4$). Arm C also learns to \emph{gate} evidence: it withholds a box on $93\%$ of non-spatial questions but supplies one on spatial questions, rather than emitting a box for everything as arm B does. The takeaway is constructive: answer-only fine-tuning silently trades away the evidence channel, but the two channels are separable, and a modest amount of grounding supervision keeps both---so the answer/evidence trade-off is a \emph{recipe} choice, not an inherent property of the model. (Full protocol, seeds, and a leakage audit in Appendix~\ref{app:sft}.)

\begin{table}[t!]
    \centering
    \caption{Four-arm fine-tuning of Qwen-2.5-VL-7B. All arms share the same answer data; B and C add the same grounding prompts, differing only in box content. Answer-only SFT (A) eliminates box emission; format rehearsal (B) restores emission but not localization; a small grounding-supervision mix (C) restores both, above every zero-shot model, while keeping the answer gain. Grounding is IoU on parseable boxes; ``fmt'' is parseable/total on the 418-box split.}
    \label{tab:sft_arms}
    \footnotesize
    \setlength{\tabcolsep}{3pt}
    \begin{tabular}{@{}lcccc@{}}
    \toprule
    & \multicolumn{3}{c}{Grounding} & SLAKE \\
    \cmidrule(r){2-4}
    Arm & fmt & IoU & Acc@0.5 & Open/Closed \\
    \midrule
    zero-shot base (7B)   & 418/418 & 0.111 & 4.1 & 37.2 / 64.6 \\
    A. answer-only        & \textbf{0/418} & -- & 0.0 & 80.0 / 77.1 \\
    B. + format rehearsal & 418/418 & 0.021 & 0.2 & 80.3 / 79.7 \\
    C. + grounding mix    & 418/418 & \textbf{0.390} & \textbf{40.2} & 83.0 / 78.4 \\
    \bottomrule
    \end{tabular}
\end{table}

\section{Limitations}
\label{sec:limitations}

\textbf{Evaluation methodology.} Our VQA scoring is a heuristic token-overlap score (synonym matching, contradiction detection, dynamic thresholds), not validated against blinded human judgment or reference-based metrics; we report it as a ``heuristic answer score'' and, where noted, alongside a normalized variant that folds plurals and radiology abbreviations symmetrically. \textbf{Grounding task design.} We use organ-level SLAKE bounding boxes with template queries, which do not capture diffuse pathologies, irregular lesions, or ambiguous boundaries where masks or point grounding would fit better. \textbf{Dataset and model scope.} Zero-shot and grounding evaluation cover VQA-RAD and SLAKE (grounding on SLAKE); the frontier panel is four API models plus two medical VLMs and one detector, and generalization to other modalities and models is unverified. \textbf{Fine-tuning scope.} The SFT study uses a single base (Qwen-2.5-VL-7B) with LoRA and full fine-tuning; the four-arm result is a controlled case study on one data mix, not a sweep over PEFT methods, mixtures, or training duration. \textbf{Causal scope.} The oracle-crop result cleanly shows that cropping to a correct box hurts VQA, since the oracle design removes localization error by construction. The further attribution of that loss specifically to \emph{context removal}, as opposed to the reduced field of view or the distribution shift induced by cropping at all, is an interpretation we do not separately test here; a control that varies the context window around a fixed oracle box (and a size-matched crop at a random location) would separate these and is the most direct follow-up to this paper.

\section{Conclusion and Future Work}

We presented a benchmark that separately evaluates the answering and grounding channels of medical VLMs, spanning four frontier models, two medical-specialized VLMs, and a dedicated detector, and asked whether grounding helps VQA.

Three findings reframe the question. First, \emph{no} off-the-shelf model localizes medical targets well---frontier, medical-specialized, and dedicated-detector systems all score mean IoU 0.05--0.24, at or barely above a trivial center-box baseline---so poor medical localization is universal, not a quirk of one model. Second, and against the ``localize-then-answer'' intuition, cropping to a bounding box degrades VQA \emph{even when the box is a perfect oracle}: closed-ended accuracy drops for every model on the matched subset, consistent with the crop discarding global context that medical VQA depends on. Grounding-by-cropping is therefore the wrong interface; better boxes would not fix it. Third, fine-tuning on answers alone silently eliminates the model's box-evidence emission, but a small amount of grounding supervision restores localization---above every zero-shot model---while preserving answer accuracy, so the answer and evidence channels are separable and jointly attainable.

The practical implication is that medical-VQA systems intended to be auditable should not rely on crop-based grounding pipelines, nor on answer-only fine-tuning that discards evidence. Promising directions are interfaces that attend to a region without discarding the rest of the image (soft attention, region prompts, or overlays rather than hard crops), medical grounding trained with explicit box or mask supervision, and evaluation that scores the evidence channel alongside the answer so that a silent loss of auditability cannot pass unnoticed.

Future work should prioritize three directions. First, and most urgently, \textbf{hybrid architectures} that decouple perception from reasoning---using specialized, domain-adapted grounding modules (e.g., Grounding DINO fine-tuned on medical data, MedSAM, or nnU-Net) to provide reliable spatial inputs to VLM reasoning---should be evaluated as a practical path to grounding-enhanced medical VQA. Second, \textbf{fine-grained failure analysis}, stratified by organ type, lesion scale, imaging modality, and anatomical laterality, is needed to build a systematic taxonomy of grounding failures and guide targeted improvements. Third, our evaluation framework should be extended with \textbf{human validation} of the automatic VQA metrics, segmentation-level grounding evaluation, and broader dataset coverage to establish a robust, community-reusable benchmark. Finally, recent advances in test-time adaptation for VLMs~\cite{singh2025ttrv} suggest that inference-time reinforcement learning could improve grounding precision without additional training data, and exploring such methods in the medical domain is a promising direction.

\appendices

\section{Model Identifiers and Access}
\label{app:models}
Frontier models are accessed through vendor APIs, so the served weights behind a
product name can change. We therefore record the exact identifier requested and the
month the runs were executed (Table~\ref{tab:model_ids}). Two labels warrant care:
the model marketed as GLM is served by the identifier \texttt{glm-4v-plus}, and the
Qwen entry in our answer comparison is the proprietary \texttt{qwen-vl-max} endpoint,
\emph{not} the open-weight Qwen2.5-VL-7B checkpoint used in the fine-tuning study.
These are different systems and we keep them distinct throughout.

\begin{table}[h]
    \centering
    \caption{API identifiers and access period for the evaluated models. All frontier
    cells are full-$N$ runs with zero empty responses, verified before use.}
    \label{tab:model_ids}
    \footnotesize
    \setlength{\tabcolsep}{3pt}
    \begin{tabular}{@{}lll@{}}
    \toprule
    Name in paper & API identifier & Accessed \\
    \midrule
    GPT-5.1 & \texttt{gpt-5.1} & 2026-07 \\
    GPT-5.5 & \texttt{openai/gpt-5.5} & 2026-07 \\
    Gemini-2.5-Pro & \texttt{gemini-2.5-pro} & 2026-07 \\
    Gemini-3-Flash & \texttt{google/gemini-3-flash-preview} & 2026-07 \\
    GLM-4V-Plus & \texttt{glm-4v-plus} & 2026-05 \\
    Qwen-VL-Max & \texttt{qwen-vl-max} & 2026-05 \\
    \midrule
    Qwen2.5-VL-7B & \texttt{Qwen/Qwen2.5-VL-7B-Instruct} & local \\
    Grounding DINO & \texttt{IDEA-Research/grounding-dino-base} & local \\
    Lingshu-7B & public release, 7B & local \\
    MedGemma-4B & public release, 4B instruct & local \\
    \bottomrule
    \end{tabular}
\end{table}

For the two medical VLMs we did not pin a checkpoint revision at run time, so we name
the released model and size rather than assert a snapshot we cannot verify after the
fact; the local weights used are preserved with the evaluation code.

\section{Laterality Confusion by Model}
\label{app:laterality}
Section~\ref{subsec:grounding} notes laterality confusion as a clinically dangerous
error mode. We quantify it on the subset of grounding queries that explicitly name a
side (e.g.\ ``Left Kidney''), $n{=}56$ per model after restricting to parseable boxes.
A \emph{side-swap} is a prediction whose centroid falls in the opposite image half from
the ground-truth centroid; because both centroids are taken in the same image
coordinates, the measure is agnostic to radiological left/right convention.

The rates (Table~\ref{tab:laterality}) span nearly the full range, which is why we
decline to state laterality confusion as a universal property: GPT-5.5 never swaps a
side, while GPT-5.1 and GLM-4V-Plus swap on roughly three quarters of lateralized
queries. Note that a low swap rate does not imply good localization---GPT-5.5 puts the
box on the correct side yet still reaches only 0.24 IoU.

\begin{table}[h]
    \centering
    \caption{Side-swap rate on lateralized grounding queries ($n{=}56$ parseable
    predictions per model).}
    \label{tab:laterality}
    \small
    \begin{tabular}{@{}lcc@{}}
    \toprule
    Model & Side-swaps & Swap rate \\
    \midrule
    GPT-5.5        & 0  & 0.0\% \\
    Gemini-2.5-Pro & 3  & 5.4\% \\
    Qwen-VL-Max    & 8  & 14.3\% \\
    Gemini-3-Flash & 18 & 32.1\% \\
    GLM-4V-Plus    & 41 & 73.2\% \\
    GPT-5.1        & 42 & 75.0\% \\
    \bottomrule
    \end{tabular}
\end{table}

\section{Fine-Tuning Protocol and Leakage Audit}
\label{app:sft}
\textbf{Arms.} All arms start from the same \texttt{Qwen2.5-VL-7B-Instruct} checkpoint
and share an identical answer-supervision set (7{,}765 examples: 5{,}972 SLAKE and
1{,}793 VQA-RAD train). Arm~A trains on that set alone. Arms~B and C additionally see
grounding-formatted prompts over the same images; they differ \emph{only} in the box
content---B receives shuffled boxes (correct format, no spatial signal), C receives the
true annotations (1{,}391 examples, about 15\% of its mix). This isolates format
rehearsal from genuine spatial supervision, which a single answer-only vs.\
grounding-mixed comparison cannot.

\textbf{Hyperparameters.} Full fine-tuning of all parameters with DeepSpeed ZeRO-3,
3 epochs, effective batch size 32 (per-device 4, gradient accumulation 8), AdamW,
learning rate $1\mathrm{e}{-5}$, cosine decay, warmup ratio 0.03. We additionally ran a
LoRA variant (rank 16, vision tower frozen) under the same data and schedule. The
box-emission collapse in arm~A reproduces at a second random seed ($0/418$ parseable
boxes in both), so it is not a single-run artifact.

\textbf{Evaluation.} The fine-tuned models are scored with the same harness, prompts,
and parser as the zero-shot models. The matched baseline is the untouched
Qwen2.5-VL-7B evaluated through that identical harness---not a differently-served API
model---so the SFT deltas are attributable to training rather than to serving
differences.

\textbf{Leakage audit.} Because SLAKE supplies both the fine-tuning data and the
grounding split, we checked image overlap directly against the training manifest:
81 of 168 grounding-evaluation images (48.2\%) also appear in the SFT training set,
while the direct-VQA test questions have 0\% overlap. We therefore report arm~C's
localization twice---over all evaluation images (IoU 0.390) and restricted to images
never seen in training (IoU 0.360, Acc@0.5 36.4)---and rely on the latter for the claim
that arm~C exceeds every zero-shot model. The two figures are close, so the effect is
not a memorization artifact.

\section{Grounding Accuracy by Target Size}
\label{app:size}
The aggregate IoU in Table~\ref{tab:grounding} hides a strong dependence on how large
the target is. Splitting the grounding split into quartiles by ground-truth box area
(Table~\ref{tab:size}) shows that performance is not uniformly poor---it \emph{collapses}
on small targets. For most models the smallest quartile scores 4--9$\times$ worse than
the largest (GPT-5.5 0.056 vs.\ 0.438; GPT-5.1 0.033 vs.\ 0.310; the untuned 7B base
0.012 vs.\ 0.261). This is the quantitative form of the scale-mismatch error mode in
Section~\ref{subsec:grounding}: asked to localize a lesion, models return a box covering
the enclosing organ, which scores near zero against a small annotation but is not
penalized when the annotation is itself large.

Two observations qualify this. First, Gemini-3-Flash is an exception: it is comparatively
flat across quartiles and is the only model that does \emph{not} peak on the largest
targets (0.243 at Q2--Q3 vs.\ 0.193 at Q4), so size dependence is a strong tendency
rather than a law. Second, grounding-supervised fine-tuning lifts the small-target
quartile the most in relative terms (0.012 $\to$ 0.237, roughly 20$\times$), i.e.\ it
does not merely inflate the easy cases. IoU here is computed end-to-end, scoring an
unparseable prediction as zero; because the frontier models emit a valid box on
99--100\% of queries, these overall values coincide with the IoU$\mid$valid figures in
Table~\ref{tab:grounding}.

\begin{table}[h]
    \centering
    \caption{Mean IoU by ground-truth target-size quartile (Q1 smallest). Quartiles are
    computed over the 418-query split.}
    \label{tab:size}
    \footnotesize
    \setlength{\tabcolsep}{2.5pt}
    \begin{tabular}{@{}lccccc@{}}
    \toprule
    Model & Q1\,(small) & Q2 & Q3 & Q4\,(large) & Overall \\
    \midrule
    GPT-5.5              & 0.056 & 0.191 & 0.285 & 0.438 & 0.243 \\
    Gemini-3-Flash       & 0.102 & 0.243 & 0.243 & 0.193 & 0.195 \\
    GPT-5.1              & 0.033 & 0.123 & 0.178 & 0.310 & 0.161 \\
    GLM-4V-Plus          & 0.026 & 0.111 & 0.141 & 0.228 & 0.126 \\
    Gemini-2.5-Pro       & 0.016 & 0.106 & 0.119 & 0.212 & 0.113 \\
    Qwen2.5-VL-7B (base) & 0.012 & 0.056 & 0.115 & 0.261 & 0.111 \\
    \midrule
    \quad + grounding mix (arm C) & 0.237 & 0.389 & 0.455 & 0.479 & 0.390 \\
    \bottomrule
    \end{tabular}
\end{table}

\bibliographystyle{IEEEtran}
\bibliography{main}

\end{document}